\documentclass{article}

\usepackage[preprint]{neurips_2026}

\usepackage[utf8]{inputenc} 
\usepackage[T1]{fontenc}    
\usepackage{hyperref}       
\usepackage{url}            
\usepackage{booktabs}       
\usepackage{amsfonts}       
\usepackage{nicefrac}       
\usepackage{microtype}      
\usepackage{xcolor}         
 \usepackage{graphicx}
 \usepackage{amsmath}
 \usepackage{float}
 \usepackage{tikz}

\usepackage{pifont}

\title{The Geometric Canary: Predicting Steerability and Detecting Drift via Representational Stability}

\author{%
  Prashant C.~Raju\\
  \texttt{rajuprashant@gmail.com}
}

\begin{document}

\maketitle

\begin{abstract}
Reliable deployment of language models requires two capabilities that appear distinct but share a common geometric foundation: predicting whether a model will accept targeted behavioral control, and detecting when its internal structure degrades. We show that geometric stability, the consistency of a representation's pairwise distance structure, addresses both. Supervised Shesha variants that measure task-aligned geometric stability predict linear steerability with near-perfect accuracy ($\rho = 0.89$--$0.97$) across 35-69 embedding models and three NLP tasks, capturing unique variance beyond class separability (partial $\rho = 0.62$--$0.76$). A critical dissociation emerges: unsupervised stability fails entirely for steering on real-world tasks ($\rho \approx 0.10$), revealing that task alignment is essential for controllability prediction. However, unsupervised stability excels at drift detection, measuring nearly $2\times$ greater geometric change than CKA during post-training alignment (up to $5.23\times$ in Llama) while providing earlier warning in 73\% of models and maintaining a $6\times$ lower false alarm rate than Procrustes. Together, supervised and unsupervised stability form complementary diagnostics for the LLM deployment lifecycle: one for pre-deployment controllability assessment, the other for post-deployment monitoring.
\end{abstract}

\section{Introduction}
\label{sec:intro}
Deploying a language model safely requires answering two questions at two different stages. Before deployment: will this model accept targeted behavioral control? After deployment: is its internal structure degrading? These questions are typically treated as separate problems, addressed by separate literatures. We show they reduce to the same underlying question: how reliable is the model's representational geometry?

\textbf{The controllability gap.} Representation engineering \citep{turner2023activation, Zou2023RepresentationEA} and activation addition \citep{subramani2022extracting, li2023inferencetime, hernandez2024linearity} have demonstrated that language models can be steered by intervening directly on internal activations. These techniques assume that concepts are encoded as stable linear directions in latent space, as formalized by the Linear Representation Hypothesis \citep{park2023the, park2025the}. But the hypothesis describes a structural property of the representation, not a guarantee: some models maintain rigid geometric structure under perturbation while others fracture. A model may classify sentiment perfectly yet have brittle representational geometry that collapses under the same steering vector that succeeds in a geometrically stable model. No existing metric predicts which models will accept steering and which will break. Classification accuracy does not suffice: a model can separate classes cleanly while encoding that separation in a fragile, perturbation-sensitive manifold.
 
\textbf{The monitoring gap.} Post-training alignment through RLHF and instruction tuning is known to reshape internal representations \citep{ouyang2022training, bai2022training}. Detecting this reorganization is critical for safety monitoring, yet the standard tools are poorly calibrated for the task. CKA \citep{kornblith2019similarity}, the most widely used similarity metric, systematically underestimates the extent of post-training geometric change because it is dominated by the top principal components and invariant to the spectral-tail reorganization that fine-tuning induces. Procrustes distance detects more change but is hypersensitive: it triggers false alarms in 44\% of cases where no functional degradation has occurred, because minor spectral perturbations accumulate in its Frobenius residual. Neither metric achieves the balance required for production monitoring, namely early detection without alarm fatigue.
 
\textbf{Our approach.} We introduce supervised variants of Shesha \citep{raju2026geometric}, a geometric stability metric that quantifies the self-consistency of a representation's pairwise distance structure. The core insight is that controllability and drift detection are both problems of geometric reliability, but they require different forms of measurement. Supervised Shesha measures whether representational geometry is aligned with task structure, which is the relevant property for predicting steering success. Unsupervised Shesha measures whether representational geometry is internally consistent regardless of task, which is the relevant property for detecting structural degradation during continuous monitoring.
 
A critical empirical dissociation validates this distinction. Unsupervised stability predicts steering success in synthetic settings where the data manifold happens to align with task structure ($\rho = 0.77$), but fails entirely on real-world NLP tasks ($\rho \approx 0.10$ on SST-2). Conversely, unsupervised stability excels at drift detection, measuring $2\times$ greater geometric change than CKA during post-training alignment while maintaining a $6\times$ lower false alarm rate than Procrustes. The two variants are not interchangeable. Each succeeds precisely where the other fails, and together they cover both phases of the deployment lifecycle.
 
\textbf{Contributions.}
\begin{enumerate}
    \item We introduce supervised Shesha variants that measure task-aligned geometric stability without requiring a trained probe.
    \item We demonstrate that supervised stability predicts linear steerability ($\rho = 0.89$--$0.97$) with unique signal beyond class separability (partial $\rho = 0.62$--$0.76$), across 69 embedding models and three NLP tasks of increasing complexity.
    \item We show that unsupervised stability detects post-training drift $2\times$ more sensitively than CKA, provides earlier warning in 73\% of models, and achieves AUC $= 0.990$ at the challenging LoRA detection boundary.
    \item We establish the stability-alignment dissociation: task-aligned and task-agnostic stability are empirically separable properties that serve complementary roles in the deployment lifecycle.
\end{enumerate}

\section{Supervised Geometric Stability}
\label{sec:shesha}

\subsection{Background}

Geometric stability quantifies the self-consistency of a representation's pairwise distance structure under internal perturbation. The core framework, Shesha \citep{raju2026geometric}, operates on Representational Dissimilarity Matrices (RDMs; \citealt{Kriegeskorte2008}). Given a representation matrix $X \in \mathbb{R}^{n \times d}$, an RDM $D \in \mathbb{R}^{n \times n}$ captures pairwise cosine distances between samples. Shesha constructs two RDMs from complementary views of the same representation and measures their agreement:
\begin{equation}
\text{Shesha}(X) = \rho_s\!\left(\text{vec}(D^{(1)}),\; \text{vec}(D^{(2)})\right)
\end{equation}
where $\rho_s$ denotes Spearman's rank correlation and $\text{vec}(\cdot)$ extracts the upper triangular elements. The choice of how to construct the two views defines distinct variants. The unsupervised \textit{feature-split} variant ($\text{Shesha}_\text{FS}$) partitions embedding dimensions into disjoint subsets, computing an RDM from each half and averaging over $K{=}30$ random splits. High values indicate that geometric structure is redundantly encoded across dimensions rather than concentrated in a few features. This variant requires no labels and is well suited for drift detection and continuous monitoring.

A key formal property distinguishes Shesha from standard similarity metrics: non-invariance to orthogonal transformations. CKA and Procrustes are invariant to rotations and reflections by construction, which means they cannot detect when fine-tuning reorganizes the spectral structure of a representation without changing its dominant subspace. Shesha retains sensitivity to these changes because feature-split RDMs depend on which dimensions encode which relational information, not merely on the overall subspace geometry. This property is what enables Shesha to detect geometric drift that similarity metrics miss (Section~\ref{sec:drift}).

\begin{figure}[t]
  \centering
  \includegraphics[width=\textwidth]{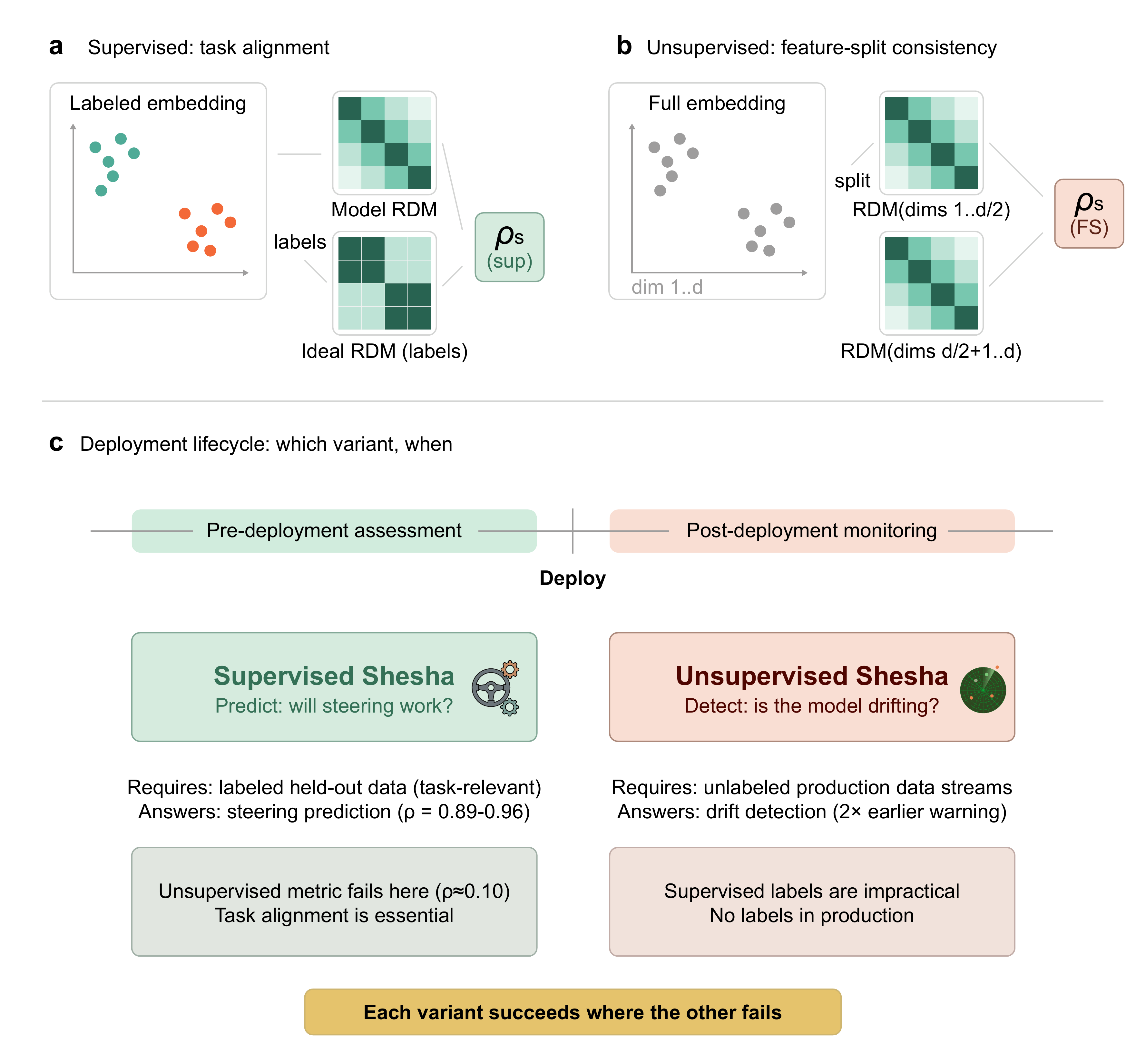}
  \caption{\footnotesize
    \textbf{Geometric stability as a deployment diagnostic: mechanism and lifecycle.} \textbf{(a)} Unsupervised Shesha ($\textrm{Shesha}_{\textrm{FS}}$) splits embedding dimensions into disjoint halves, computes a representational dissimilarity matrix (RDM) from each half, and measures their rank correlation. High values indicate that pairwise distance structure is redundantly encoded across features. No labels are required. \textbf{(b)} Supervised Shesha ($\textrm{Shesha}_{\textrm{sup}}$) correlates the model's RDM with an ideal RDM derived from task labels, measuring how faithfully representational geometry reflects class structure. \textbf{(c)} The two variants serve complementary phases of the deployment lifecycle. Pre-deployment, supervised Shesha predicts whether a model will accept linear steering ($\rho = 0.89-0.97$). Post-deployment, unsupervised Shesha detects geometric drift ($2\times$ more sensitive than CKA). The cross marks highlight the dissociation: unsupervised stability fails for steering prediction ($\rho \approx 0.10$ on SST-2), while supervised stability is impractical for continuous monitoring without labels. Each variant succeeds where the other fails.%
  }
  \label{fig:rc-texture}
\end{figure}

\subsection{Task-aligned variants}

The unsupervised variant answers whether a representation's geometry is internally consistent. For controllability prediction, the relevant question is different: is the geometry aligned with task structure? We introduce four supervised variants that incorporate label information $y \in \{1, \ldots, C\}^n$.

\textbf{Supervised RDM alignment} ($\text{Shesha}_\text{sup}$) directly correlates the model's RDM with an ideal label-derived RDM:
\begin{equation}
\text{Shesha}_\text{sup}(X, y) = \rho_s\!\left(\text{vec}(D_X),\; \text{vec}(D_y)\right)
\end{equation}
where $D_y$ encodes label dissimilarity (Hamming distance on one-hot encodings). This measures how faithfully the representation's distance structure reflects task-relevant categories, without training a probe.

\textbf{Variance ratio} ($\text{Shesha}_\text{var}$) provides a computationally efficient approximation:
\begin{equation}
\text{Shesha}_\text{var}(X, y) = \frac{\sum_{c=1}^{C} n_c \|\mu_c - \mu\|^2}{\sum_{i=1}^{n} \|x_i - \mu\|^2}
\end{equation}
where $\mu_c$ is the centroid of class $c$, $\mu$ is the global mean, and $n_c$ is the number of samples in class $c$. This captures the proportion of total variance attributable to class structure.

\textbf{Class separation ratio} ($\text{Shesha}_\text{sep}$) operates in distance space rather than variance space:
\begin{equation}
\text{Shesha}_\text{sep}(X, y) = \frac{\bar{d}_\text{between}}{\bar{d}_\text{within}}
\end{equation}
where $\bar{d}_\text{between}$ and $\bar{d}_\text{within}$ are mean pairwise distances between and within classes, respectively, estimated via bootstrap subsampling ($B{=}50$ iterations, 50\% subsampling rate). This variant is related to Fisher's discriminant ratio but operates on pairwise distances rather than projected variances.

\textbf{LDA subspace stability} ($\text{Shesha}_\text{LDA}$) measures the consistency of the optimal linear decision boundary under resampling:
\begin{equation}
\text{Shesha}_\text{LDA}(X, y) = \frac{1}{B} \sum_{b=1}^{B} \left| \hat{w}^\top \hat{w}^{(b)} \right|
\end{equation}
where $\hat{w}$ is the unit-normalized LDA discriminant direction fitted on the full dataset and $\hat{w}^{(b)}$ is the direction fitted on bootstrap subsample $b$. High values indicate that the discriminant subspace is robust to sampling variation; low values suggest the linear separability is an artifact of the particular sample rather than a stable geometric property.

\subsection{The key distinction}

These variants address two fundamentally different questions. Unsupervised $\text{Shesha}_\text{FS}$ asks: is the geometry internally consistent? Supervised variants ask: is the geometry aligned with a specific task? Internal consistency and task alignment are logically independent properties. A representation can be internally rigid yet poorly organized for a given task, or well organized for a task yet fragile under feature perturbation. The experiments in Sections~3 and~4 demonstrate that this logical independence is also empirical: unsupervised stability fails to predict steering success on real-world tasks (Section~3.3) yet excels at detecting post-training drift (Section~4). The two flavors of stability serve complementary diagnostic roles precisely because they measure different things.

\section{Supervised Stability Predicts Steering Performance}
\label{sec:steering}
If a language model's representational geometry determines whether steering interventions succeed or fail, then geometric stability should predict controllability before any steering is attempted. We test this hypothesis across three settings of increasing complexity.

\begin{figure*}[t]
\centering
\includegraphics[width=\textwidth]{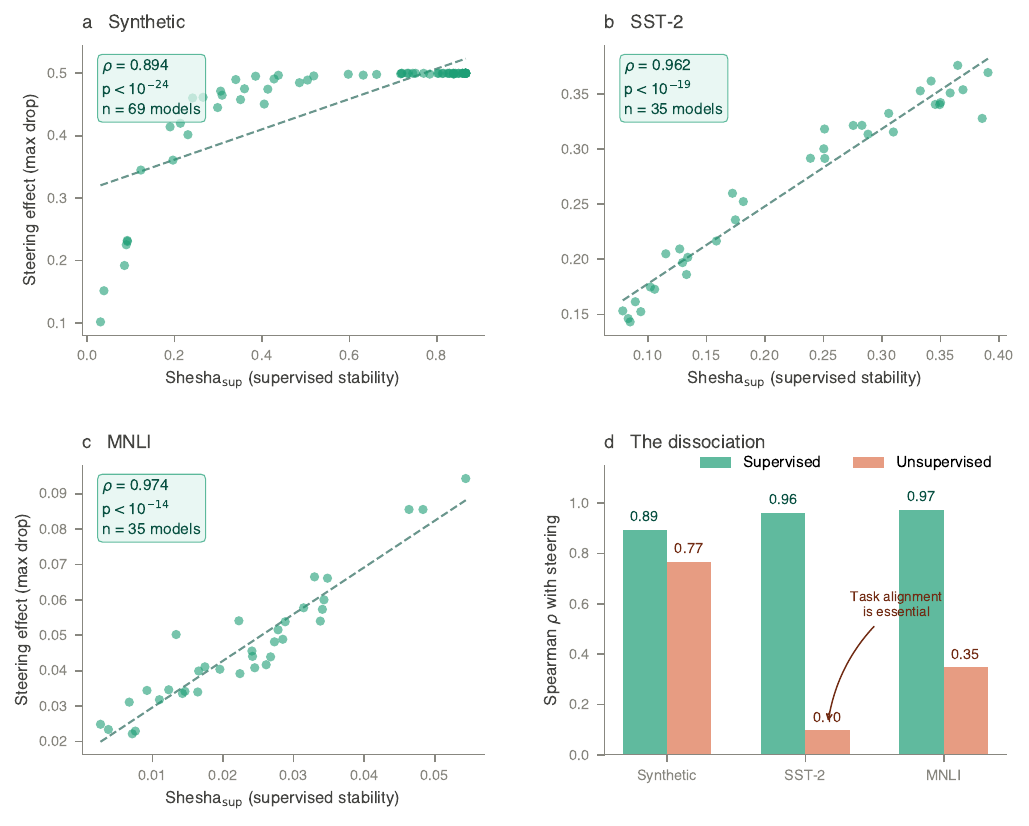}
\caption{\textbf{Supervised geometric stability predicts linear steerability across all settings.}
\textbf{(a--c)} Scatter plots of supervised Shesha (computed on held-out Set~A) versus steering effectiveness (max accuracy drop, evaluated on disjoint Set~B) for each model, averaged across 15 random seeds.
\textbf{(a)}~Synthetic sentiment (69 models, $\rho = 0.894$, $p < 10^{-24}$).
\textbf{(b)}~SST-2 binary sentiment (35 models, $\rho = 0.962$, $p < 10^{-20}$).
\textbf{(c)}~MNLI ternary NLI (35 models, $\rho = 0.974$, $p < 10^{-22}$).
Dashed lines show linear fits.
\textbf{(d)}~The stability-alignment dissociation. Supervised Shesha (teal) maintains high correlation with steering across all settings, while unsupervised feature-split Shesha (coral) collapses from $\rho = 0.77$ on synthetic data to $\rho \approx 0.10$ on SST-2. In synthetic settings, the data manifold aligns with task structure by construction, so generic rigidity suffices. On real-world tasks, task-aligned measurement is essential for predicting controllability.}
\label{fig:steering}
\end{figure*}

\subsection{Experimental design}

\textbf{Models.} Experiment~1 (Synthetic) evaluated 69 sentence embedding models spanning 11 architecture families (MiniLM, DistilBERT, MPNet, BERT, RoBERTa, DeBERTa, E5, BGE, GTE, UAE, SimCSE) across three size tiers and both supervised contrastive and unsupervised pretraining objectives. Experiments~2 and~3 evaluated a subset of 35 models on SST-2 \citep{socher-etal-2013-recursive} (binary sentiment) and MNLI \citep{williams-etal-2018-broad} (ternary natural language inference), respectively.

\textbf{Data and split-half protocol.} The synthetic dataset consists of 1,000 sentiment-laden sentences generated from a combinatorial grammar ($4 \times 8 \times 8 = 256$ unique sentences per polarity), designed to avoid lexical memorization effects. SST-2 and MNLI use their standard benchmarks. In all settings, data is partitioned into two completely disjoint sets: Set~A ($n = 500$ synthetic; $n = 400$ SST-2/MNLI) for metric computation and Set~B for steering evaluation (split further into training and testing). This strict separation ensures that no information leaks between metric estimation and steering assessment. All experiments use 15 random seeds, yielding 1,035 observations (Synthetic) and 525 each (SST-2, MNLI).

\textbf{Steering protocol.} For each model and seed: (1) train a logistic regression probe on the Set~B training split; (2) extract the weight vector $w$ as the steering direction (for MNLI, the top right singular vector of the coefficient matrix $W = U\Sigma V^\top$, capturing the principal axis of class discrimination); (3) compute steered embeddings $e' = e + \alpha \hat{w}$ for $\alpha \in \{-2, -1.5, \ldots, 1.5, 2\}$; (4) evaluate probe accuracy on the Set~B test split; (5) record the maximum accuracy drop: $\text{max\_drop} = \text{acc}_0 - \min_\alpha \text{acc}(\alpha)$.

\textbf{Metrics.} On Set~A, we compute supervised Shesha ($\text{Shesha}_\text{sup}$), unsupervised feature-split Shesha ($\text{Shesha}_\text{FS}$), Fisher discriminant, silhouette score, Procrustes alignment, and anisotropy.

\textbf{Negative controls.} Two controls validate the methodology: (1) shuffled labels, which recompute all supervised metrics with permuted class assignments; (2) random directions, which average the maximum accuracy drop over 20 random unit vectors per split.

\subsection{Supervised stability predicts controllability}


\textbf{Unique signal beyond separability.} The raw correlations are comparable to those of the Fisher discriminant ($\rho = 0.888$, $0.885$, $0.952$). To test whether stability captures something beyond class separation, we computed partial correlations controlling for both the Fisher discriminant and silhouette score. Supervised Shesha maintained large partial correlations across all settings: $\rho_\text{partial} = 0.665$ (Synthetic), $0.764$ (SST-2), and $0.620$ (MNLI), all $p < 0.001$. By contrast, unsupervised Shesha retained no residual signal after the same controls ($\rho_\text{partial} < 0.10$). This establishes that the reliability of class structure under perturbation is a distinct predictive factor that static separability metrics miss. Separability may be necessary for steering, but stability is what ensures control.

\textbf{Negative controls.} Shuffled-label controls confirm that supervised metrics capture genuine task structure: $\text{Shesha}_\text{sup}$ drops to $-0.001$ under label permutation across all three settings ($p < 10^{-10}$). Random-direction controls quantify the signal-to-noise ratio of true steering vectors: true directions produce $10.8\times$ (Synthetic), $2.7\times$ (SST-2), and $1.3\times$ (MNLI) larger accuracy drops than random directions. The declining ratio reflects narrowing control margins as task complexity increases, yet Shesha continues to identify steerable models even when the margin for intervention is slim.

\subsection{The stability-alignment dissociation}

The most consequential finding is a dissociation between supervised and unsupervised stability. Unsupervised feature-split Shesha correlates with steering in the synthetic setting ($\rho = 0.77$, $p < 10^{-14}$), where the data manifold happens to be well aligned with the sentiment axis. On real-world tasks, this correlation collapses: $\rho = 0.10$ on SST-2 (n.s.) and $\rho = 0.35$ on MNLI (n.s.). After controlling for separability, the residual drops below $\rho_\text{partial} = 0.10$ in all settings (Figure~\ref{fig:steering}D).

This dissociation has a clear interpretation. In synthetic data, the combinatorial grammar constructs a manifold whose principal axes of variation coincide with sentiment polarity. Generic geometric rigidity therefore happens to align with the task. In naturalistic settings, task-relevant structure occupies a low-dimensional subspace of a much richer manifold: a model can be geometrically rigid overall yet poorly organized for any particular downstream task. Intrinsic consistency is not the same as task alignment.

\textbf{Training objectives shape steerability.} Model rankings are consistent across settings: supervised contrastive models (BGE, E5, GTE families) are the most steerable, while unsupervised variants (unsup-SimCSE, E5-base-unsupervised) and retrieval-specialized models (multi-qa-*) are the least steerable. Supervised contrastive training produces both the class separation and the geometric rigidity required for reliable linear intervention.

\textbf{Practical implication.} These results establish $\text{Shesha}_\text{sup}$ as an a priori diagnostic for controllability. Given a model and a labeled calibration set, a practitioner can compute supervised stability on held-out data and predict whether the model will accept steering, without running any steering experiments. Models with high supervised stability will steer reliably; models with low stability will fracture under intervention regardless of classification accuracy.

\section{Stability Detects Representational Drift}
\label{sec:drift}
Section~\ref{sec:steering} established that geometric stability is a prerequisite for controllability. This raises the stakes: if stability is what makes control possible, then detecting when it degrades becomes urgent. But supervised stability requires labeled data and a known task, which makes it impractical for continuous production monitoring. Post-deployment surveillance needs an unsupervised metric. This is precisely where the unsupervised variant, which failed for steering prediction, finds its purpose. We evaluate drift detection across four experiments that progressively isolate the properties of an ideal monitoring metric: sensitivity to real geometric change (Experiment~1), monotonic response to controlled perturbations (Experiment~2), predictive validity for downstream task performance (Experiment~3), and calibrated false alarm rates (Experiment~4).

\begin{figure*}[t]
\centering
\includegraphics[width=\textwidth]{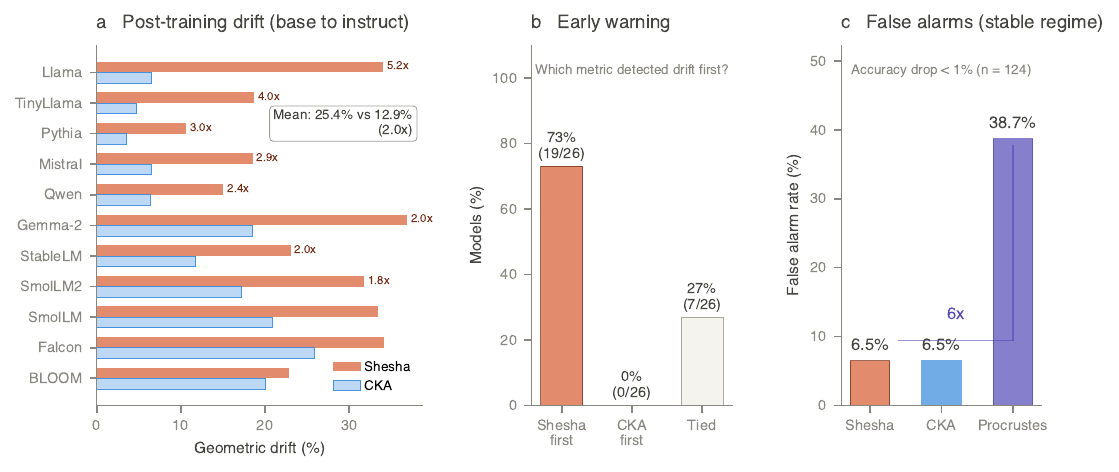}
\caption{\textbf{Unsupervised Shesha detects drift earlier than CKA while avoiding Procrustes' false alarms.}
\textbf{(a)}~Post-training geometric drift between 23 base/instruct model pairs spanning 11 families (0.14B--7B parameters), averaged across four prompt types. Shesha detects $1.96\times$ greater drift than CKA on average, with family-specific ratios ranging from $1.1\times$ (BLOOM) to $5.2\times$ (Llama), indicating distributed geometric reorganization that CKA's top-principal-component weighting underestimates.
\textbf{(b)}~Early warning analysis on 26 sentence embedding models under progressive Gaussian noise (51 levels, $\sigma \in [0, 0.5]$). Using a 5\% detection threshold, Shesha detected drift first in 73\% of models; CKA never detected first. When metrics diverged, Shesha won 100\% of the time.
\textbf{(c)}~False alarm rates in the stable regime (accuracy drop $< 1\%$) on LoRA perturbation benchmarks ($n = 124$ observations). Procrustes triggers false alarms in 38.7\% of stable cases compared to 6.5\% for Shesha, a $6\times$ difference. This oversensitivity stems from spectral-tail noise accumulating in Procrustes' Frobenius residual. Shesha achieves the optimal balance: more sensitive than CKA, more specific than Procrustes.}
\label{fig:canary}
\end{figure*}

\subsection{Shesha detects greater geometric change than CKA}

\textbf{Experiment 1: Post-training drift.} We compared representations from 23 base/instruct model pairs spanning 11 families (Qwen, Llama, SmolLM, SmolLM2, Mistral, StableLM, Gemma, TinyLlama, Pythia, BLOOM, Falcon) ranging from 0.14B to 7B parameters. For each pair, we embedded four semantically distinct prompt sets (factual, descriptive, instructions, conversational; 50 prompts each) and computed drift as $1 - \rho_s(\text{RDM}_\text{base}, \text{RDM}_\text{instruct})$ for Shesha, with analogous dissimilarity formulations for CKA and Procrustes. Embeddings were extracted from the final layer using mean pooling and L2 normalization.

Shesha detected nearly $2\times$ greater geometric change than CKA on average (25.1\% vs.\ 12.9\%, ratio: $1.96\times$). The discrepancy varied substantially across families (Figure~\ref{fig:canary}A): Llama showed the largest gap ($5.23\times$; 34.0\% Shesha vs.\ 6.5\% CKA), while BLOOM ($1.14\times$) and Falcon ($1.32\times$) showed near-parity. Prompt type also modulated the ratio: factual ($2.37\times$) and descriptive ($2.28\times$) prompts induced the largest discrepancies, while instruction prompts showed the smallest ($1.44\times$), suggesting that instruction tuning optimizes representational geometry specifically for instruction-following inputs while producing broader reorganization for out-of-distribution prompts.

\textbf{Experiment 2: Structured perturbations.} To characterize metric response curves under controlled conditions, we applied three perturbation types to 16 causal LMs: Gaussian noise ($\sigma \in [0.01, 0.50]$), quantization (INT8, INT4), and LoRA modifications (rank 1--64, initialization scale 0.001--0.1). All metrics increased monotonically with perturbation magnitude. At high noise ($\sigma = 0.5$), Shesha captured $1.7\times$ more drift than CKA (71\% vs.\ 43\%). Quantization effects were modest: INT8 induced 0.74\% Shesha drift with negligible accuracy loss, while INT4 induced 3.6\% drift, a form of geometric compression that preserves functional topology. LoRA drift scaled with both rank and initialization magnitude, with large perturbations (rank 64 or init scale 0.1) causing 5--18\% accuracy drops.

\subsection{All metrics predict functional degradation equally}

Greater sensitivity is only useful if it detects real problems. To validate that geometric drift is a reliable proxy for functional degradation, we ran canary experiments that inject progressive noise and track both drift and downstream accuracy.

\textbf{Experiment 3: Canary validation.} We injected Gaussian noise at 51 levels ($\sigma \in [0.00, 0.50]$) into 26 sentence embedding models and measured both geometric drift and SST-2 classification accuracy (5-fold CV, logistic regression). All three primary metrics achieved nearly identical Spearman correlations with accuracy drop: Shesha ($\rho = 0.927$), CKA ($\rho = 0.937$), and Procrustes ($\rho = 0.935$). This equivalence replicated on 15 causal language models under Gaussian noise, quantization, and LoRA perturbations: Shesha ($\rho = 0.915$), CKA ($\rho = 0.912$), Procrustes ($\rho = 0.903$). The consistently high correlations confirm that geometric drift reliably predicts functional degradation regardless of which metric is chosen. The metrics differ not in \textit{what} they predict, but in \textit{when} and \textit{how reliably} they detect it.

\subsection{Shesha provides earlier warning than CKA}

Using a 5\% detection threshold on the canary validation, Shesha provided earlier warning in 73\% of models (19/26), while CKA detected earlier in 0\%, with 27\% tied. When the metrics diverged, Shesha detected first 100\% of the time. Mean detection thresholds confirmed the gap: Shesha triggered at $\sigma = 0.123$ versus CKA at $\sigma = 0.136$. To validate this advantage independent of threshold selection, we performed ROC analysis on the LoRA perturbation benchmark, where structural drift is subtle and detection is most challenging. Shesha achieved the highest AUC (0.990) compared to Procrustes (0.988) and CKA (0.987). At the operationally relevant 5\% false positive rate, Shesha maintained 90.2\% sensitivity, while Procrustes dropped to 85.4\%. This confirms that Shesha's earlier detection reflects genuine signal rather than threshold artifacts.

\subsection{Procrustes triggers excessive false alarms}

While Procrustes achieves high predictive validity and detects drift earliest in absolute terms ($\sigma = 0.040$), this sensitivity comes at a severe cost. In the stable regime (accuracy drop $< 1\%$), Procrustes triggered false alarms in 44\% of cases compared to 7.3\% for both Shesha and CKA, a $6\times$ difference (Figure~\ref{fig:canary}C). At minimal perturbation where functional performance was unchanged, Procrustes reported 1.50\% drift versus 0.04\% for Shesha, a $37\times$ inflation. The mechanism is spectral. Procrustes minimizes the Frobenius residual $\|XR - Y\|_F$ over the full spectrum after optimal rotation. In high-dimensional representations, small perturbations accumulate in the spectral tail as noise that Procrustes attempts to align but cannot, inflating the distance score. CKA avoids this by upweighting dominant eigenvalues, but at the cost of underestimating distributed geometric change. Shesha's rank-based correlation occupies the middle ground: it is robust to low-magnitude perturbations that preserve relative distance ordering (avoiding false alarms) while remaining sensitive to reorganization that disrupts pairwise relationships (enabling early warning). 

\textbf{Practical recommendation.} For production monitoring, Shesha serves as the primary metric: sensitive enough to detect geometric fracturing before functional failure, yet specific enough to avoid the alarm fatigue that undermines operator confidence. CKA provides stable confirmation as a conservative baseline. Procrustes should be reserved for settings where false alarms are acceptable, such as security-critical deployments where any geometric change warrants investigation.

\section{Discussion}
\label{sec:discussion}

\subsection{Two phases, one principle}

The results in Sections~\ref{sec:steering} and~\ref{sec:drift} are two acts of a single story. Before deployment, a practitioner computes supervised Shesha on task-relevant held-out data to determine whether a model's representational geometry will support linear steering interventions. After deployment, the same practitioner monitors unsupervised Shesha continuously to detect geometric reorganization before it manifests as task failure. These are not competing diagnostics but complementary phases of the same geometric principle, deployed where each variant excels.

\subsection{Why the distinction explains steering vs. drift}

The central empirical finding of this paper is a dissociation: unsupervised stability fails to predict steering ($\rho \approx 0.10$ on SST-2) yet excels at detecting drift ($2\times$ more sensitive than CKA). This is not a paradox. It follows directly from what each variant measures.

Steering is a \textit{task-directed} intervention. A steering vector is computed relative to a specific semantic axis (e.g., sentiment polarity), and its success depends on whether the representation organizes information along that axis in a geometrically consistent way. A model can have perfectly rigid overall geometry while encoding the task-relevant subspace in a fragile or disorganized manner. Unsupervised Shesha, which measures generic feature-split consistency across the entire embedding space, cannot distinguish between rigidity that happens to align with the task and rigidity that is orthogonal to it. This is why unsupervised stability predicts steering in synthetic settings, where the data manifold's principal axes coincide with sentiment polarity by construction, but collapses on naturalistic tasks where task-relevant structure occupies a small subspace of a much richer manifold. Supervised Shesha succeeds because it directly measures whether the geometry respects task structure.

Drift detection operates under fundamentally different conditions. When monitoring a model for post-training degradation, the question is not whether the geometry is aligned with any particular task, but whether the geometry has changed at all. This is an intrinsic, task-agnostic question: has the pairwise distance structure between inputs been reorganized? Unsupervised Shesha is ideally suited for this because it measures exactly this property, the self-consistency of representational geometry, without requiring labeled data or a predefined task. The $2\times$ sensitivity advantage over CKA arises because CKA's invariance to orthogonal transformations blinds it to spectral-tail reorganization, the very kind of distributed geometric change that instruction tuning induces.


This explanatory framework yields a clean boundary condition. Unsupervised stability suffices when the relevant question is intrinsic to the data manifold: has the geometry changed (drift detection), or is the geometry internally coherent (biological perturbation analysis, compression detection)? Supervised stability is required when the relevant question involves alignment with an external semantic structure: will the model accept task-directed control (steering), or does the geometry support downstream transfer? In prior work \citep{raju2026geometric}, unsupervised stability similarly fails for transfer prediction ($\rho = 0.03$) while succeeding at detecting compression-induced manifold damage. The dissociation is not a limitation of the framework; it is its most informative feature, telling practitioners exactly which variant to deploy and when.

\subsection{Implications for representation engineering}

The Linear Representation Hypothesis \citep{park2023the, park2025the} posits that concepts are encoded as linear directions in activation space, providing the theoretical foundation for steering and activation engineering. Our results show that this hypothesis carries an implicit assumption: that the linear structure is geometrically stable. Supervised Shesha makes this assumption testable and quantitative. Models from supervised contrastive training (BGE, E5, GTE) consistently exhibit both the class separation and the geometric rigidity required for reliable linear intervention. Unsupervised and retrieval-specialized models do not, regardless of their classification accuracy. This suggests that the success of representation engineering depends not just on the existence of linear structure but on its robustness to perturbation, a property that current evaluation practices do not assess.

\subsection{Implications for safety monitoring}

Current post-deployment monitoring relies primarily on behavioral probes that detect failure after it has occurred: a drop in downstream accuracy, an increase in toxic outputs, or a shift in calibration. Geometric monitoring offers an earlier signal by detecting the structural preconditions for failure before behavioral symptoms appear. The $5.23\times$ discrepancy between Shesha and CKA in Llama suggests that instruction tuning causes substantial manifold reorganization that existing representational tools miss entirely. At the same time, the low false alarm rate (7.3\% versus Procrustes' 44\%) is critical for operational adoption: a monitoring metric that cries wolf loses the trust of the operators it is designed to protect.

\subsection{Limitations and future work}

Several boundary conditions apply. Our steering experiments test sentence embedding models rather than decoder hidden states during generation; extending to token-level stability in autoregressive models is an important next step. The drift analysis uses held-out prompts rather than the training distribution itself, and supervised Shesha requires labeled data for metric computation (though not for model training). Computationally, split-half estimation requires multiple forward passes, and the metric operates globally rather than isolating localized geometric damage in specific subspaces. These limitations suggest natural extensions: token-level stability profiles for decoder models, layer-wise analysis to identify where drift concentrates within the network, online monitoring systems with adaptive thresholds, and connections to mechanistic interpretability. If supervised stability identifies which subspaces support linear control, and mechanistic interpretability identifies which subspaces correspond to identifiable circuits, their intersection may reveal which circuits are robust and which are fragile.

\begin{ack}
We thank Padma K. and Annapoorna Raju for generously supporting the computational resources used in this work. We thank the many institutions and individuals whose open-source datasets, frameworks, and models were used in our work. The authors acknowledge the use of large language models (specifically the GPT, Claude, and Gemini families) to assist with (limited) code factoring, debugging, and text polishing. All hypotheses, experimental designs, analyses, and interpretations were independently formulated and verified by the authors, and the authors assume full responsibility for all content and claims in this work.
\end{ack}

\section*{Code Availability}
\label{sec:code}
The full code necessary to reproduce all experiments, benchmarks, and analysis described in this paper is publicly available at \url{https://github.com/prashantcraju/geometric-canary}.


\bibliographystyle{apalike}  
\bibliography{references} 

\newpage
\appendix

\vbox{%
\hsize\textwidth
\linewidth\hsize
\vskip 0.1in
\centering
{\LARGE\bf
Appendix
\par}
\vskip 0.29in
  \vskip -\parskip
  \hrule height 1pt
  \vskip 0.09in%
}

\section*{Contents}

\label{app:appendix}
\begin{itemize}
    \item Appendix~\ref{app:related-work}: Related Work
    \item Appendix~\ref{app:steering}: Steering: Extended Methods and Results
    \item Appendix~\ref{app:drift-detailed}: Representational Drift Detection: Extended Methods and Results
    \item Appendix~\ref{app:controls}: Negative Control Details

\end{itemize}

\newpage


\section{Related Work} 
\label{app:related-work}
\textbf{Representation similarity metrics.} Several methods exist for comparing neural network representations, including Centered Kernel Alignment \citep[CKA;][]{kornblith2019similarity}, Singular Vector Canonical Correlation Analysis \citep[SVCCA;][]{raghu2017svcca}, Projection Weighted CCA \citep[PWCCA;][]{morcos2018insights}, Representational Similarity Analysis \citep[RSA;][]{Kriegeskorte2008}, and Procrustes distance~\citep{Schnemann1966,Rohlf1990,Masarotto2018,dryden1998statistical}. These methods measure external similarity: whether two representations encode similar pairwise structures or occupy aligned subspaces. Shesha addresses a fundamentally different question, internal stability, asking whether a single representation's geometry is consistent under feature subsampling or data perturbation rather than whether two representations are similar. This distinction matters because similarity and stability are empirically dissociable \citep{raju2026geometric, shesha2026}: representations can be highly similar yet geometrically fragile, or stable yet dissimilar.

\textbf{Geometric properties of representations.} Prior work has characterized static geometric properties of representations, including intrinsic dimensionality \citep{ansuini2019intrinsic}, linear decodability \citep{Cohen2020, Recanatesi2021}, hierarchical organization \citep{park2025the}, and topological structure \citep{naitzat2020topology}. Recent work has shown that representation dispersion, the average pairwise cosine distance among hidden vectors, predicts downstream accuracy and perplexity across model families without requiring labeled data~\citep{li2026dispersion}. These approaches describe what the geometry looks like but do not assess whether it is reliable across data splits, sampling noise, or distribution shift. Shesha complements these static characterizations by measuring the consistency of geometric structure under perturbation; critically, dispersion and stability decouple under compression, as high-performing models can exhibit broad embedding geometry while remaining geometrically fragile under feature subsampling, a dissociation we previously termed the ``geometric tax'' \citep{raju2026geometric}.

\textbf{Representation steering and control.} Language models can be controlled by intervening on internal activations, a technique termed activation engineering or steering \citep{turner2023activation, subramani2022extracting}. Representation Engineering \citep{Zou2023RepresentationEA} and causal interventions \citep{meng2022locating, geiger2024finding} rely on the Linear Representation Hypothesis \citep{park2023the, park2025the}, which posits that concepts are encoded as stable linear directions in latent space. While these methods exploit geometric structure for behavioral control, they do not quantify the stability of that structure. If representational geometry is brittle, steering vectors may become unreliable across contexts or model updates. Our work provides the first metric that predicts steering success a priori by measuring the geometric stability that these techniques implicitly assume.
 
\textbf{Representational drift and robustness.} Representational drift has been studied during fine-tuning \citep{kumar2022finetuning, aghajanyan2020intrinsic}, under adversarial perturbations \citep{ilyas2019adversarial}, and across random seeds \citep{nguyen2021do}. Model stitching \citep{bansal2021revisiting} and cross-architecture comparisons assess feature compatibility but lack a unified framework for quantifying geometric stability. Work on RLHF and preference optimization has focused primarily on behavioral drift \citep{ouyang2022training, bai2022training}, though recent studies have begun to identify geometric degradation during alignment \citep{Springer2026Geometry}. Our drift detection experiments provide the first systematic comparison of geometric metrics for post-alignment monitoring, demonstrating that standard tools either underestimate change (CKA) or trigger excessive false alarms (Procrustes).
 
\textbf{Alignment and capability tradeoffs.} RLHF and related alignment methods are known to induce an alignment tax on broader capabilities \citep{ouyang2022training, bai2022training}. Studies have identified representation collapse under preference optimization \citep{moalla2024no} and mode collapse in RLHF \citep{omahony2024mode} through behavioral metrics. Geometric characterization of these effects remains limited. Our analysis quantifies the geometric reorganization that instruction tuning induces across 11 model families, revealing up to $5.23\times$ greater change than CKA detects and establishing unsupervised Shesha as a monitoring metric calibrated for this regime.

\newpage


\section{Steering: Extended Methods and Results} 
\label{app:steering}
 \subsection{Model lists}
 
\textbf{Experiment 1 (Synthetic).} We evaluated 69 sentence embedding models spanning 11 architecture families: MiniLM (L3, L6, L12), DistilBERT, MPNet, BERT, RoBERTa, DeBERTa, E5 (small, base, large, multilingual), BGE (small, base, large), GTE (small, base, large), UAE, and multiple SimCSE variants (supervised and unsupervised). Models covered three size tiers (small: $<$30M, base: 30--120M, large: $>$120M parameters) and included both supervised contrastive and unsupervised/pretrained objectives.
 
\textbf{Experiments 2--3 (SST-2, MNLI).} 35 sentence embedding models from the MiniLM, MPNet, DistilBERT, BGE, E5, GTE, and SimCSE families. All models were evaluated using mean-pooled outputs from the base encoder for consistency.
 
\subsection{Synthetic data generation}
 
We generated a corpus of 1,000 sentiment-laden sentences using a combinatorial
grammar that was generated by a large language model (Claude Sonnet) to ensure broad geometric coverage:
\begin{verbatim}
template: "{context}, the {noun} was {adjective}"
contexts = ["in my opinion", "overall",
            "considering everything", "to be honest"]
nouns = ["aspect", "element", "part", "feature",
         "component", "unit", "item", "factor"]
adj_pos = ["adequate", "fine", "good", "decent",
           "solid", "excellent", "superb", "exceptional"]
adj_neg = ["poor", "bad", "mediocre", "lacking",
           "subpar", "terrible", "awful", "dreadful"]
\end{verbatim}
This produces $4 \times 8 \times 8 = 256$ unique positive and 256 unique negative sentences per polarity (1,000 total with resampling). The combinatorial structure avoids lexical memorization effects while ensuring broad geometric coverage.
 
\subsection{Split-half protocol}
 
For each seed, data was partitioned into completely disjoint sets: Set~A ($n = 500$ synthetic; $n = 400$ SST-2/MNLI) for metric computation and Set~B for steering evaluation (split further into equal training and testing subsets). This strict separation ensures that Shesha computed on Set~A cannot trivially predict steering on Set~B through shared samples. All experiments used 15 random seeds: $\{3, 7, 9, 11, 12, 18, 103, 108, 320, 411, 724, 1754, 1991, 2222, 7258\}$.
 
\subsection{Full metric correlation tables}
 
Table~\ref{tab:steering_synthetic} reports all metric correlations with steering effectiveness for Experiment~1 (Synthetic).
 
\begin{table}[H]
\centering
\caption{Experiment 1: Metric correlations with steering (Synthetic). Spearman $\rho$ between geometric metrics and max drop, aggregated by model ($n = 69$). Partial correlations control for Fisher discriminant and silhouette score.}
\label{tab:steering_synthetic}
\small
\begin{tabular}{llccc}
\toprule
\textbf{Category} & \textbf{Metric} & \textbf{Raw $\rho$} & $\boldsymbol{p}$ & \textbf{Partial $\rho$} \\
\midrule
Stability & Shesha (Supervised) & 0.894 & $< 10^{-24}$ & 0.665 \\
          & Shesha (Unsupervised) & 0.767 & $< 10^{-14}$ & 0.053 \\
          & Procrustes & 0.797 & $< 10^{-16}$ & $-$0.089 \\
Separability & Fisher Discriminant & 0.888 & $< 10^{-24}$ & -- \\
             & Silhouette Score & 0.889 & $< 10^{-24}$ & -- \\
Structure & Anisotropy & 0.710 & $< 10^{-11}$ & -- \\
\bottomrule
\end{tabular}
\end{table}
 
\begin{table}[H]
\centering
\caption{Experiment 2: Metric correlations with steering (SST-2, $n = 35$ models).}
\label{tab:steering_sst2}
\small
\begin{tabular}{llccc}
\toprule
\textbf{Category} & \textbf{Metric} & \textbf{Raw $\rho$} & $\boldsymbol{p}$ & \textbf{Partial $\rho$} \\
\midrule
Stability & Shesha (Supervised) & 0.962 & $< 10^{-19}$ & 0.764 \\
          & Shesha (Unsupervised) & 0.103 & n.s. & $-$0.036 \\
          & Procrustes & 0.976 & $< 10^{-23}$ & 0.522 \\
Separability & Fisher Discriminant & 0.885 & $< 10^{-12}$ & -- \\
             & Silhouette Score & 0.884 & $< 10^{-12}$ & -- \\
Structure & Anisotropy & 0.396 & 0.019 & -- \\
\bottomrule
\end{tabular}
\end{table}
 
\begin{table}[H]
\centering
\caption{Experiment 3: Metric correlations with steering (MNLI, $n = 35$ models).}
\label{tab:steering_mnli}
\small
\begin{tabular}{llccc}
\toprule
\textbf{Category} & \textbf{Metric} & \textbf{Raw $\rho$} & $\boldsymbol{p}$ & \textbf{Partial $\rho$} \\
\midrule
Stability & Shesha (Supervised) & 0.974 & $< 10^{-22}$ & 0.620 \\
          & Shesha (Unsupervised) & 0.221 & n.s. & 0.100 \\
          & Procrustes & 0.886 & $< 10^{-12}$ & 0.491 \\
Separability & Fisher Discriminant & 0.952 & $< 10^{-18}$ & -- \\
             & Silhouette Score & 0.650 & $< 10^{-5}$ & -- \\
Structure & Anisotropy & 0.343 & 0.044 & -- \\
\bottomrule
\end{tabular}
\end{table}
 
\begin{table}[H]
\centering
\caption{Summary: Supervised Shesha predicts steering across all settings.}
\label{tab:steering_summary}
\small
\begin{tabular}{lccccl}
\toprule
\textbf{Dataset} & \textbf{Models} & \textbf{Obs.} & \textbf{Shesha (Raw)} & \textbf{Shesha (Partial)} & \textbf{Verdict} \\
\midrule
Synthetic & 69 & 1,035 & 0.894 & 0.665*** & Primary predictor \\
SST-2 & 35 & 525 & 0.962 & 0.764*** & Independent signal \\
MNLI & 35 & 525 & 0.974 & 0.620*** & Task invariant \\
\bottomrule
\multicolumn{6}{l}{\footnotesize ***$p < 0.001$} \\
\end{tabular}
\end{table}
 
\subsection{Model ranking analysis}
 
Consistent patterns emerged across all three settings. The most steerable models were supervised contrastive embeddings: BGE-large, GTE-large, E5-large, and UAE-large consistently appeared in the top quartile. These models are trained with explicit pairwise supervision that produces both class separation and geometric rigidity.
 
The least steerable models fell into two categories: (1) unsupervised variants (unsup-SimCSE, E5-base-unsupervised), which lack the task-aligned geometric structure that steering requires, and (2) retrieval-specialized models (multi-qa-MiniLM, multi-qa-MPNet), which optimize for query-document matching rather than class-level geometric organization. Notably, retrieval models often achieve competitive classification accuracy yet fail under steering, confirming that separability alone is insufficient for controllability.

\newpage


\section{Representational Drift Detection: Extended Methods and Results}
\label{app:drift-detailed}
\subsection{Model Lists}
\label{app:drift-models}

\begin{table}[H]
\centering
\caption{Base/Instruct model pairs for post-training drift analysis (Experiment 1).}
\label{tab:drift-pairs}
\begin{tabular}{llc}
\toprule
\textbf{Base Model} & \textbf{Instruct Model} & \textbf{Params} \\
\midrule
HuggingFaceTB/SmolLM-135M & SmolLM-135M-Instruct & 0.14B \\
HuggingFaceTB/SmolLM2-135M & SmolLM2-135M-Instruct & 0.14B \\
HuggingFaceTB/SmolLM-360M & SmolLM-360M-Instruct & 0.36B \\
HuggingFaceTB/SmolLM2-360M & SmolLM2-360M-Instruct & 0.36B \\
Qwen/Qwen2-0.5B & Qwen2-0.5B-Instruct & 0.5B \\
Qwen/Qwen1.5-0.5B & Qwen1.5-0.5B-Chat & 0.5B \\
bigscience/bloom-560m & bloomz-560m & 0.56B \\
EleutherAI/pythia-1b & pythia-1b-deduped & 1.0B \\
meta-llama/Llama-3.2-1B & Llama-3.2-1B-Instruct & 1.0B \\
TinyLlama/TinyLlama-1.1B-intermediate-step-1431k-3T & TinyLlama-1.1B-Chat-v1.0 & 1.1B \\
bigscience/bloom-1b1 & bloomz-1b1 & 1.1B \\
Qwen/Qwen2-1.5B & Qwen2-1.5B-Instruct & 1.5B \\
stabilityai/stablelm-2-1\_6b & stablelm-2-zephyr-1\_6b & 1.6B \\
HuggingFaceTB/SmolLM-1.7B & SmolLM-1.7B-Instruct & 1.7B \\
HuggingFaceTB/SmolLM2-1.7B & SmolLM2-1.7B-Instruct & 1.7B \\
Qwen/Qwen1.5-1.8B & Qwen1.5-1.8B-Chat & 1.8B \\
google/gemma-2b & gemma-2b-it & 2.0B \\
google/gemma-2-2b & gemma-2-2b-it & 2.0B \\
meta-llama/Llama-3.2-3B & Llama-3.2-3B-Instruct & 3.0B \\
Qwen/Qwen1.5-4B & Qwen1.5-4B-Chat & 4.0B \\
Qwen/Qwen2-7B & Qwen2-7B-Instruct & 7.0B \\
mistralai/Mistral-7B-v0.1 & Mistral-7B-Instruct-v0.1 & 7.0B \\
tiiuae/falcon-7b & falcon-7b-instruct & 7.0B \\
\bottomrule
\end{tabular}
\end{table}

\begin{table}[H]
\centering
\caption{Causal language models for structured perturbation analysis (Experiment 2, 16 models) and extended canary validation (Experiment 4, 15 models). Experiment 4 excludes SmolLM2-1.7B due to SST-2 evaluation constraints. Both experiments applied Gaussian noise, quantization, and LoRA perturbations.}
\label{tab:causal-models}
\begin{tabular}{llcc}
\toprule
\textbf{Model} & \textbf{Family} & \textbf{Params} & \textbf{Exp 4} \\
\midrule
HuggingFaceTB/SmolLM-135M & SmolLM & 0.14B & \checkmark \\
HuggingFaceTB/SmolLM2-135M & SmolLM2 & 0.14B & \checkmark \\
HuggingFaceTB/SmolLM-360M & SmolLM & 0.36B & \checkmark \\
HuggingFaceTB/SmolLM2-360M & SmolLM2 & 0.36B & \checkmark \\
Qwen/Qwen2-0.5B & Qwen2 & 0.5B & \checkmark \\
meta-llama/Llama-3.2-1B & Llama & 1.0B & \checkmark \\
TinyLlama/TinyLlama-1.1B-intermediate-step-1431k-3T & TinyLlama & 1.1B & \checkmark \\
Qwen/Qwen2-1.5B & Qwen2 & 1.5B & \checkmark \\
stabilityai/stablelm-2-1\_6b & StableLM & 1.6B & \checkmark \\
HuggingFaceTB/SmolLM-1.7B & SmolLM & 1.7B & \checkmark \\
HuggingFaceTB/SmolLM2-1.7B & SmolLM2 & 1.7B & No \\
google/gemma-2b & Gemma & 2.0B & \checkmark \\
google/gemma-2-2b & Gemma-2 & 2.0B & \checkmark \\
meta-llama/Llama-3.2-3B & Llama & 3.0B & \checkmark \\
Qwen/Qwen2-7B & Qwen2 & 7.0B & \checkmark \\
mistralai/Mistral-7B-v0.1 & Mistral & 7.0B & \checkmark \\
\bottomrule
\end{tabular}
\end{table}

\begin{table}[H]
\centering
\caption{Sentence embedding models for canary validation (Experiment 3, 26 models).}
\label{tab:canary-models}
\begin{tabular}{ll}
\toprule
\textbf{Model} & \textbf{Family} \\
\midrule
sentence-transformers/paraphrase-MiniLM-L3-v2 & MiniLM \\
sentence-transformers/all-MiniLM-L6-v2 & MiniLM \\
sentence-transformers/paraphrase-MiniLM-L6-v2 & MiniLM \\
sentence-transformers/all-MiniLM-L12-v2 & MiniLM \\
sentence-transformers/multi-qa-MiniLM-L6-cos-v1 & MiniLM \\
sentence-transformers/all-mpnet-base-v2 & MPNet \\
sentence-transformers/paraphrase-mpnet-base-v2 & MPNet \\
sentence-transformers/multi-qa-mpnet-base-cos-v1 & MPNet \\
sentence-transformers/distilbert-base-nli-mean-tokens & DistilBERT \\
sentence-transformers/all-distilroberta-v1 & DistilRoBERTa \\
sentence-transformers/paraphrase-distilroberta-base-v1 & DistilRoBERTa \\
sentence-transformers/bert-base-nli-mean-tokens & BERT \\
sentence-transformers/stsb-roberta-base & RoBERTa \\
sentence-transformers/nli-roberta-base-v2 & RoBERTa \\
sentence-transformers/paraphrase-albert-small-v2 & ALBERT \\
thenlper/gte-small & GTE \\
thenlper/gte-base & GTE \\
thenlper/gte-large & GTE \\
intfloat/e5-small-v2 & E5 \\
intfloat/e5-base-v2 & E5 \\
intfloat/e5-large-v2 & E5 \\
BAAI/bge-small-en-v1.5 & BGE \\
BAAI/bge-base-en-v1.5 & BGE \\
BAAI/bge-large-en-v1.5 & BGE \\
princeton-nlp/sup-simcse-bert-base-uncased & SimCSE \\
princeton-nlp/unsup-simcse-bert-base-uncased & SimCSE \\
\bottomrule
\end{tabular}
\end{table}

\subsection{Prompt sets}
 
For each model pair, we embedded four semantically coherent prompt sets (50 prompts each, 200 total per pair):
\begin{itemize}
    \item \textbf{Factual:} Declarative statements about scientific facts (e.g., ``The Earth orbits around the Sun.'')
    \item \textbf{Descriptive:} Vivid scene descriptions (e.g., ``Waves crashed against the rocky shoreline.'')
    \item \textbf{Instructions:} Explanatory requests (e.g., ``Explain how photosynthesis works in plants.'')
    \item \textbf{Conversational:} Casual dialogue prompts (e.g., ``How was your day today?'')
\end{itemize}
All prompts were generated synthetically using large language models (Claude Sonnet, Gemini, ChatGPT). This was done to ensure that each prompt was semantically coherent within its category while maintaining diversity. The complete prompt sets are available in our code repository.
 
\subsection{Embedding extraction}
 
For all models, hidden states were extracted from the final layer using mean pooling over non-padding tokens, followed by L2 normalization. Chat templates were applied where available (detected via \texttt{tokenizer.chat\_template}), with \texttt{add\_generation\_prompt=False} to avoid assistant prefix bias. Maximum sequence length was 256 tokens.
 
\subsection{Drift metrics}
 
All drift metrics are reported as dissimilarity (higher = more drift):
\begin{itemize}
    \item \textbf{Shesha}: $1 - \rho_{\text{Spearman}}(\text{RDM}_{\text{clean}}, \text{RDM}_{\text{noisy}})$
    \item \textbf{RDM-Pearson}: $1 - r_{\text{Pearson}}(\text{RDM}_{\text{clean}}, \text{RDM}_{\text{noisy}})$
    \item \textbf{CKA}: $1 - \text{CKA}_{\text{debiased}}$
    \item \textbf{Procrustes}: $1 - \text{Procrustes similarity}$ after optimal orthogonal alignment
    \item \textbf{Wasserstein}: Sliced Wasserstein distance with 100 random projections
\end{itemize}
 
\subsection{Drift by prompt type}
 
The Shesha/CKA ratio varied systematically across prompt types: factual ($2.37\times$), descriptive ($2.28\times$), conversational ($1.82\times$), and instructions ($1.44\times$). The lower ratio for instruction prompts suggests that instruction tuning optimizes representational geometry specifically for the instruction-following distribution, producing less distributed reorganization for in-distribution inputs and greater reorganization for out-of-distribution prompts.
 
\subsection{Canary validation: predictive validity}
 
In the canary validation on 15 causal language models (0.14B--7B parameters) under Gaussian noise ($\sigma \in [0.00, 0.50]$), all primary metrics achieved strong predictive validity for downstream accuracy degradation: Shesha ($\rho = 0.915$), CKA ($\rho = 0.912$), and Procrustes ($\rho = 0.903$). The consistently high correlations confirm that geometric drift reliably predicts functional degradation regardless of the metric chosen. The metrics differ not in what they predict, but in when and how reliably they detect it.
 
\subsection{ROC analysis}
 
On the LoRA perturbation benchmark, where structural drift is subtle and detection is most challenging, all metrics achieved high overall performance: Shesha (AUC = 0.990), Procrustes (AUC = 0.988), and CKA (AUC = 0.987). At the operationally relevant 5\% false positive rate, Shesha maintained 90.2\% sensitivity compared to 85.4\% for Procrustes, confirming that Shesha's earlier detection reflects genuine signal rather than threshold artifacts.
 
\subsection{False alarm mechanism}
 
Procrustes' oversensitivity stems from its spectral properties. Procrustes minimizes the Frobenius residual $\|XR - Y\|_F$ over the full spectrum after optimal rotation. In high-dimensional representations, small perturbations accumulate in the spectral tail as noise that Procrustes attempts to align but cannot, inflating the distance score. At minimal perturbation (LoRA init scale = 0.001), where accuracy was essentially unchanged ($\Delta\text{Acc} \approx 0.08\%$), Procrustes registered $37\times$ more drift than Shesha (1.50\% vs. 0.04\%). CKA avoids this by upweighting dominant eigenvalues, but at the cost of underestimating distributed change. Shesha's rank-based correlation provides the middle ground: robust to low-magnitude perturbations that preserve relative distance ordering, yet sensitive to reorganization that disrupts pairwise relationships.

\subsection{Results}
\begin{table}[H]
\centering
\caption{Geometric drift between base and instruction-tuned model pairs. Drift metrics (as percentages) averaged across four prompt types (factual, descriptive, instructions, conversational). Higher values indicate greater representational change from instruction tuning.}
\label{tab:base-instruct-comparison}
\begin{tabular}{llccc}
\toprule
\textbf{Model Pair} & \textbf{Size} & \textbf{Shesha (\%)} & \textbf{CKA (\%)} & \textbf{Procrustes (\%)} \\
\midrule
SmolLM-135M $\to$ Instruct & 135M & 28.3 & 18.3 & 13.4 \\
SmolLM2-135M $\to$ Instruct & 135M & 29.6 & 12.8 & 10.0 \\
SmolLM-360M $\to$ Instruct & 360M & 41.6 & 24.5 & 17.5 \\
SmolLM2-360M $\to$ Instruct & 360M & 30.5 & 19.1 & 14.1 \\
Qwen2-0.5B $\to$ Instruct & 0.5B & 10.9 & 4.9 & 2.9 \\
Qwen1.5-0.5B $\to$ Chat & 0.5B & 14.2 & 6.0 & 4.4 \\
BLOOM-560M $\to$ BLOOMZ & 560M & 35.5 & 31.6 & 23.1 \\
Pythia-1B $\to$ Deduped & 1.0B & 10.6 & 3.5 & 2.6 \\
Llama-3.2-1B $\to$ Instruct & 1.0B & 34.9 & 6.7 & 4.2 \\
TinyLlama-1.1B $\to$ Chat & 1.1B & 18.7 & 4.7 & 3.1 \\
BLOOM-1.1B $\to$ BLOOMZ & 1.1B & 10.3 & 8.6 & 5.8 \\
Qwen2-1.5B $\to$ Instruct & 1.5B & 13.0 & 5.0 & 22.3 \\
StableLM-1.6B $\to$ Zephyr & 1.6B & 23.0 & 11.8 & 19.4 \\
SmolLM-1.7B $\to$ Instruct & 1.7B & 30.4 & 19.6 & 26.2 \\
SmolLM2-1.7B $\to$ Instruct & 1.7B & 35.2 & 19.6 & 28.0 \\
Qwen1.5-1.8B $\to$ Chat & 1.8B & 21.1 & 7.2 & 15.1 \\
Gemma-2B $\to$ IT & 2.0B & 32.4 & 14.6 & 22.0 \\
Gemma-2-2B $\to$ IT & 2.0B & 41.4 & 22.5 & 25.6 \\
Llama-3.2-3B $\to$ Instruct & 3.0B & 33.1 & 6.3 & 13.8 \\
Qwen1.5-4B $\to$ Chat & 4.0B & 15.5 & 6.4 & 13.8 \\
Qwen2-7B $\to$ Instruct & 7.0B & 15.6 & 8.6 & 17.5 \\
Mistral-7B $\to$ Instruct & 7.0B & 18.5 & 6.5 & 14.1 \\
Falcon-7B $\to$ Instruct & 7.0B & 34.1 & 25.8 & 26.8 \\
\bottomrule
\end{tabular}
\end{table}

\begin{table}[H]
\centering
\caption{Mean drift by Gaussian noise level (averaged across 16 models and 4 prompt types).}
\label{tab:gaussian-results}
\begin{tabular}{lccc}
\toprule
\textbf{Noise Level ($\sigma$)} & \textbf{Shesha} & \textbf{CKA} & \textbf{Procrustes} \\
\midrule
0.01 & 0.003 & 0.002 & 0.019 \\
0.02 & 0.012 & 0.007 & 0.037 \\
0.05 & 0.049 & 0.032 & 0.085 \\
0.10 & 0.119 & 0.074 & 0.141 \\
0.15 & 0.225 & 0.146 & 0.206 \\
0.20 & 0.361 & 0.238 & 0.278 \\
0.30 & 0.630 & 0.380 & 0.362 \\
0.40 & 0.714 & 0.432 & 0.396 \\
0.50 & 0.716 & 0.430 & 0.414 \\
\bottomrule
\end{tabular}
\end{table}

\subparagraph{Quantization.}
Table~\ref{tab:quant-results} shows drift induced by quantization. INT8 quantization caused minimal representational drift (Shesha: 2.1\%, CKA: 1.4\%, Procrustes: 5.3\%), while INT4 quantization induced approximately 3$\times$ more drift (Shesha: 6.2\%, CKA: 4.2\%, Procrustes: 9.7\%). Procrustes showed the highest sensitivity to quantization effects. However, given the minimal functional degradation observed in Experiment 4 under quantization, this heightened sensitivity may largely reflect rigid geometric rotations rather than functional erosion.

\begin{table}[H]
\centering
\caption{Mean drift by quantization level (averaged across 16 models and 4 prompt types).}
\label{tab:quant-results}
\begin{tabular}{lccc}
\toprule
\textbf{Quantization} & \textbf{Shesha} & \textbf{CKA} & \textbf{Procrustes} \\
\midrule
INT8 & 0.021 & 0.014 & 0.053 \\
INT4 (NF4) & 0.062 & 0.042 & 0.097 \\
\bottomrule
\end{tabular}
\end{table}

\subparagraph{LoRA.}
Table~\ref{tab:lora-rank-results} shows drift as a function of LoRA rank at a fixed initialization scale (0.01). Drift increased monotonically with rank, from near-zero at $r = 1$ to substantial reorganization at $r = 64$ (Shesha: 15.3\%, CKA: 8.8\%, Procrustes: 13.9\%). Table~\ref{tab:lora-scale-results} shows drift as a function of initialization scale at a fixed rank ($r = 8$). The initialization scale had a dramatic effect: increasing from 0.001 to 0.1 increased Shesha drift from 0.06\% to 44.2\%, demonstrating that the magnitude of LoRA perturbation, not just the rank, determines representational impact.

\begin{table}[H]
\centering
\caption{Mean drift by LoRA rank at fixed initialization scale (0.01).}
\label{tab:lora-rank-results}
\begin{tabular}{lccc}
\toprule
\textbf{LoRA Rank} & \textbf{Shesha} & \textbf{CKA} & \textbf{Procrustes} \\
\midrule
1 & 0.002 & 0.001 & 0.017 \\
2 & 0.004 & 0.002 & 0.022 \\
4 & 0.047 & 0.035 & 0.055 \\
8 & 0.029 & 0.016 & 0.048 \\
16 & 0.064 & 0.036 & 0.075 \\
32 & 0.086 & 0.052 & 0.102 \\
64 & 0.153 & 0.088 & 0.139 \\
\bottomrule
\end{tabular}
\end{table}

\begin{table}[H]
\centering
\caption{Mean drift by LoRA initialization scale at fixed rank ($r = 8$).}
\label{tab:lora-scale-results}
\begin{tabular}{lccc}
\toprule
\textbf{Init Scale} & \textbf{Shesha} & \textbf{CKA} & \textbf{Procrustes} \\
\midrule
0.001 & 0.001 & 0.000 & 0.010 \\
0.005 & 0.006 & 0.003 & 0.023 \\
0.01 & 0.029 & 0.015 & 0.047 \\
0.02 & 0.095 & 0.050 & 0.106 \\
0.05 & 0.303 & 0.176 & 0.237 \\
0.10 & 0.441 & 0.324 & 0.394 \\
\bottomrule
\end{tabular}
\end{table}

\begin{table}[H]
\centering
\caption{Gaussian noise perturbation on sentence embedding models (26 models, STS-B task). Drift metrics show strong correlation with accuracy drop ($\rho > 0.92$ for all metrics).}
\label{tab:canary-sentence-embeddings}
\begin{tabular}{lcccc}
\toprule
\textbf{Noise ($\sigma$)} & \textbf{Accuracy Drop (\%)} & \textbf{Shesha} & \textbf{CKA} & \textbf{Procrustes} \\
\midrule
0.00 & 0.0 & 0.000 & 0.000 & 0.000 \\
0.01 & 0.0 & 0.001 & 0.000 & 0.015 \\
0.02 & 0.0 & 0.002 & 0.002 & 0.031 \\
0.03 & 0.0 & 0.005 & 0.003 & 0.046 \\
0.04 & 0.0 & 0.008 & 0.006 & 0.061 \\
0.05 & 0.0 & 0.012 & 0.009 & 0.076 \\
0.06 & 0.2 & 0.019 & 0.014 & 0.092 \\
0.07 & 0.1 & 0.025 & 0.018 & 0.107 \\
0.08 & 0.3 & 0.033 & 0.024 & 0.122 \\
0.09 & 0.3 & 0.042 & 0.031 & 0.138 \\
0.10 & 0.4 & 0.054 & 0.039 & 0.154 \\
\vdots \\
0.15 & 1.4 & 0.115 & 0.090 & 0.232 \\
0.20 & 2.8 & 0.208 & 0.171 & 0.315 \\
0.25 & 5.0 & 0.329 & 0.271 & 0.394 \\
0.30 & 9.4 & 0.488 & 0.411 & 0.484 \\
0.35 & 12.9 & 0.603 & 0.543 & 0.565 \\
0.40 & 16.9 & 0.705 & 0.657 & 0.631 \\
0.45 & 21.0 & 0.771 & 0.728 & 0.683 \\
0.50 & 22.5 & 0.813 & 0.772 & 0.716 \\
\bottomrule
\end{tabular}
\end{table}

\newpage

\section{Negative Control Details}
\label{app:controls}
\subsection{Shuffled label controls}
 
To confirm that supervised metrics reflect genuine task structure rather than spurious geometric patterns, we recomputed all supervised metrics with randomly permuted class labels. Supervised Shesha collapsed to $-0.001$ under label shuffling across all three settings:
\begin{itemize}
    \item Synthetic: $0.60 \to -0.001$ ($p < 10^{-10}$)
    \item SST-2: $0.23 \to -0.001$ ($p < 10^{-10}$)
    \item MNLI: $0.02 \to -0.001$ ($p < 10^{-10}$)
\end{itemize}
The complete collapse confirms that the metric captures task-relevant structure. The decreasing baseline values (0.60, 0.23, 0.02) reflect increasing geometric complexity: in synthetic data, the manifold is strongly organized by sentiment; in MNLI, the ternary structure is more diffuse.
 
\subsection{Random direction controls}
 
To quantify the signal-to-noise ratio of true steering directions, we compared the maximum accuracy drop from the true probe direction against the average drop from 20 random unit vectors per split:
\begin{itemize}
    \item Synthetic: $10.8\times$ larger effect ($p < 10^{-10}$)
    \item SST-2: $2.7\times$ larger effect ($p < 10^{-10}$)
    \item MNLI: $1.3\times$ larger effect
\end{itemize}
The declining ratio reflects narrowing steering margins as task complexity increases: MNLI's ternary classification distributes class structure more diffusely across the manifold, leaving less room for large directional effects. Crucially, Shesha continues to identify steerable models even when the margin for intervention is slim, because the metric captures the reliability of whatever class structure exists rather than its magnitude.
 
\subsection{Early warning: per-model breakdown}
 
Among the 26 sentence embedding models in the canary validation, Shesha detected drift before CKA in 19 models (73\%). The 7 tied cases corresponded to models where both metrics crossed the 5\% threshold at the same noise level ($\sigma = 0.10$ or $\sigma = 0.15$), typically models with moderate dimensionality (384--768 dimensions) where the spectral gap between dominant and tail components is small. In no case did CKA detect earlier than Shesha. Mean detection thresholds: Shesha $\sigma = 0.123$, CKA $\sigma = 0.136$.
 
 \newpage

\end{document}